\documentclass{article}

     \usepackage[preprint, nonatbib]{neurips_2019}

\usepackage[utf8]{inputenc} 
\usepackage[T1]{fontenc}    
\usepackage{hyperref}       
\usepackage{url}            
\usepackage{booktabs}       
\usepackage{amsfonts}       
\usepackage{nicefrac}       
\usepackage{microtype}      
\usepackage{graphicx}
\usepackage{wrapfig}
\usepackage{chngcntr}
\usepackage{float}
\usepackage{lipsum}
\usepackage{subfig}

\title{Push it to the Limit: Discover Edge-Cases in Image Data with Autoencoders}

\author{
  Ilja Manakov \\
  Ludwig Maximilian University\\
  Munich, Germany \\
  \texttt{ilja.manakov@med.uni-muenchen.de} \\
  \And
  Volker Tresp \\
  Ludwig Maximilian University, Siemens AG\\
  Munich, Germany \\
  \texttt{volker.tresp@siemens.com}
}

\begin{document}

\maketitle

\begin{abstract}
In this paper, we focus on the problem of identifying semantic factors of variation in large image datasets.
By training a convolutional Autoencoder on the image data, we create encodings, which describe each datapoint at a higher level of abstraction than pixel-space.
We then apply Principal Component Analysis to the encodings to disentangle the factors of variation in the data.
Sorting the dataset according to the values of individual principal components, we find that samples at the high and low ends of the distribution often share specific semantic characteristics.
We refer to these groups of samples as semantic groups.
When applied to real-world data, this method can help discover unwanted edge-cases. 

\end{abstract}
\section{Introduction}
Deep Neural Networks are incredibly hungry for data. Typical training datasets consist of at least several thousands of samples.
Such datasets are often either generated automatically by relying on heuristics (e.g., scraping from the internet) or collected over time from established processes (e.g., as part of the clinical workflow).
Validating such datasets is crucial for avoiding biases and incorrect samples, yet, as their size grows, this task quickly becomes prohibitively expensive.
Image data is particularly hard to validate, since not all relevant factors, such as image quality, are readily available, which necessitates visual inspection of the data.
\begin{wrapfigure}[17]{r}{0.4\textwidth}
	\includegraphics[width=\linewidth]{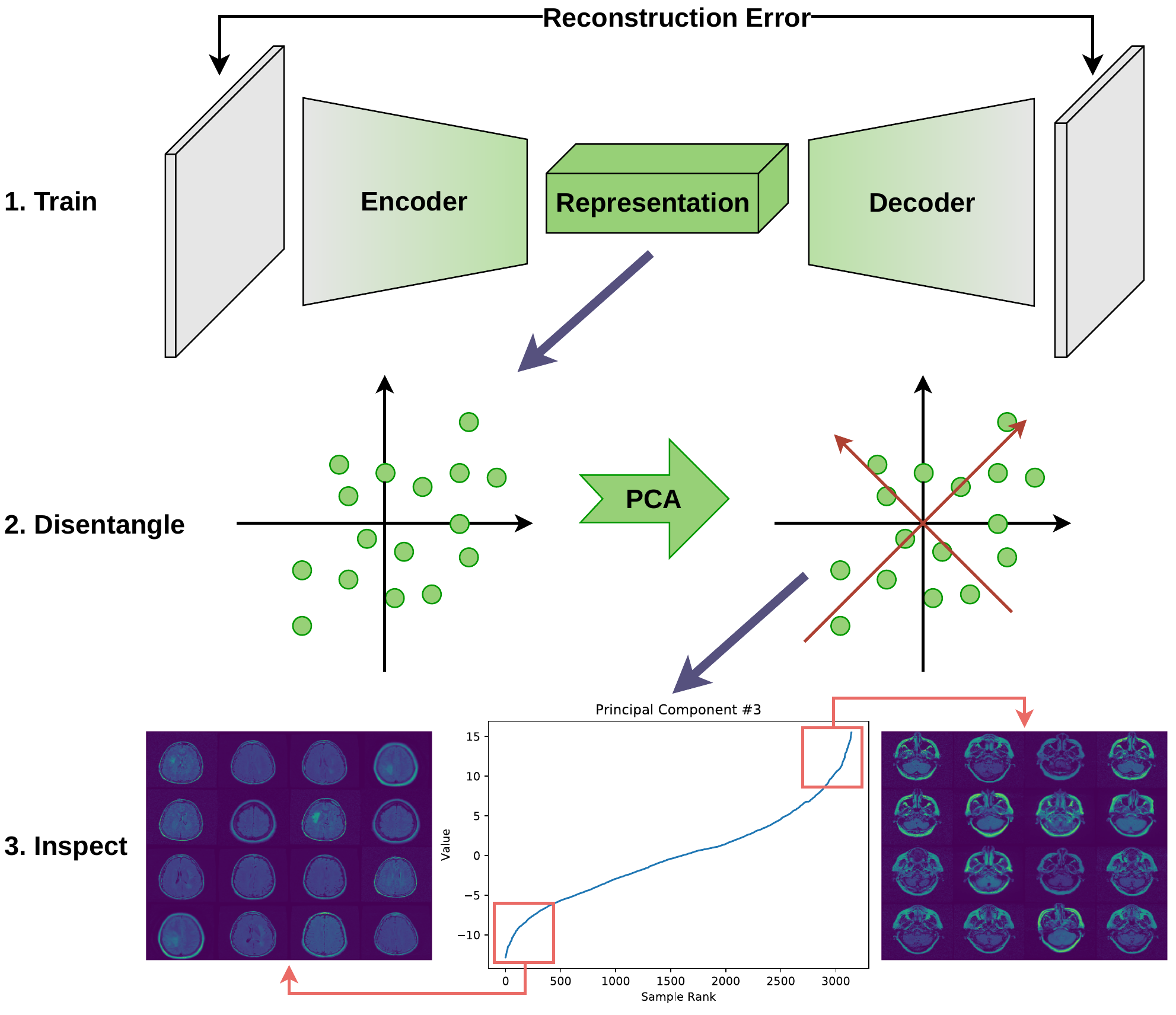}
	\caption{Schematic of the process: First, we train an Autoencoder.
	Then, we apply PCA to disentangle the representations.
	Finally, inspect the dataset by sorting the samples according to the principal component values.} \label{process}
\end{wrapfigure}
Existing methods that address this issue fall generally into two categories.
The first category attempts to provide tools for easier data exploration, such as data visualization and image retrieval methods \cite{gordo16, gillmann17}.
These promote investigation at the individual sample level.
The other one encompasses methods like active learning \cite{settles10}, outlier detection \cite{hyunha15} and identifying unknown unknowns \cite{lakkaraju17, attenberg15}.
These methods offer assistance in a dataset level investigation but require the training of a supervised predictive model on the dataset.
Our approach falls closer to the second category, as it enables dataset level investigation.
However it does not require any labeled data and is thus generically applicable to all image data.
\section{Methods}
To summarize, our approach consists of three steps.
The first step is to train an unsupervised model on the data to obtain high-level representations of the samples in the dataset.
In the second step, we disentangle the semantic factors in the representations using Principal Component Analysis (PCA).
In the final step, we sort the dataset based on each principal component.
Looking at samples with the highest and smallest values for each component reveals the semantic groups in the data.
We demonstrate the entire process in figure Fig. \ref{process}.

We used a deep convolutional Autoencoder as the unsupervised model, due to its simplicity and known ability to learn expressive representations \cite{girdhar16, hou19}.
An Autoencoder is a network that consists of two parts: an encoder and a decoder \cite{rumelhart86}.
It compresses an input sample to a latent representation with reduced dimensionality via the encoder and then restores it to full size via the decoder.
The rationale is that the network has to describe the datapoint at a higher level of abstraction to preserve the most amount of information possible through the bottleneck that is the latent representation.
The reconstruction error between the original sample and the reconstruction from the Autoencoder constitutes the loss signal in this scenario.
Since we are working with image data, the layers in our encoder and decoder consist of convolutions, as they provide a good prior for this type of data \cite{ulyanov18}.

We then used the trained model to convert the dataset into a collection of representations by passing the samples through the Autoencoder and extracting the resulting features at the last layer of the encoder.
As the Autoencoder representations are highly distributed, changing a single dimension in the representation results in a negligible change in the reconstruction.
In an attempt to disentangle the semantic factors that underlie the image generating process,  we applied PCA \cite{wold87} to the representations.
PCA is a statistical method, which identifies the directions of the greatest variance in the data.
All directions found by PCA are pairwise orthogonal and ordered in a descending fashion by the amount of variance they account for.

Finally, we used the principal components to inspect the dataset by sorting the samples according to the value of the principal component under study.
In this way, we found that semantically distinct groups of samples form for extreme values of some principal components.
\section{Experiments}
We tested our methodology on three different imaging modalities: retinal Optical Coherence Tomography (OCT), cranial Magnetic Resonance Imaging (MRI) and chest radiographs (CXR).
All datasets are publicly available on Kaggle.
In all our experiments, we used a kernel size of three for the convolutions in the Autoencoder.
For downsampling, we used strided convolutions and for upsampling, we used bilinear interpolation followed by padded convolutions \cite{odena16}.
After each down-/upsampling layer, we placed two residual blocks \cite{he16} with two padded convolutions each for increased model complexity.
We doubled/halved the number of channels with every down-/upsampling layer and chose the number of layers such that the resulting representation would have a total size of 8192 neurons.
The full implementation and training details can be found online at [REDACTED].
Additionally, we dedicated 20\% of the data for validation to check whether the Autoencoder would overfit to the training data.
In all cases, we did not experience overfitting, as the validation error did not rise during training (see Fig. \ref{losses} in the Appendix).
We trained all models on a single NVidia GeForce 1070Ti.
\paragraph{OCT}
For the first experiment, we used the AMD SD-OCT dataset from Farsiu et al. \cite{farsiu14}, which contains a total of 38400 OCT b-scans (i.e., 2D retinal depth profiles).
Each b-scan has a height of 512 pixels and a width of 1000 pixels.
We used reflective padding to increase the image width to 1024 to obtain a final shape of 512 x 1024 pixels, which resulted in a representation of shape 256 x 4 x 8.
We trained the model for 300 epochs, which took roughly three days.
\paragraph{MRI}
We performed our second experiment on the brain MRI dataset from Buda et al. \cite{buda19}.
It consists of 2633 images, showing axial brain MRI scans. Each image is the combination of pre-contrast, FLAIR, and post-contrast sequences as its three channels.
For the training of the Autoencoder, we used every sequence individually, resulting in a total number of 7899 grayscale images.
All images are 256 x 256 pixels in size, which led to a representation of size 128 x 8 x 8.
The Autoencoder trained for 1450 epochs in about a day.
\paragraph{CXR}
In our final experiment, we applied our methodology to the pneumonia chest x-ray dataset from Kermany et al. \cite{kermany18}.
This dataset encompasses 5863 chest x-ray images.
The images have varying aspect ratios and sizes.
For simplicity, we resized all of them to 256 x 256 pixels and trained our network for 1300 epochs ($\sim$ one day) on the resized images.
\begin{wrapfigure}[34]{hr}{0.4\textwidth}
	\centering
	\subfloat[MRI]{\includegraphics[width=\linewidth]{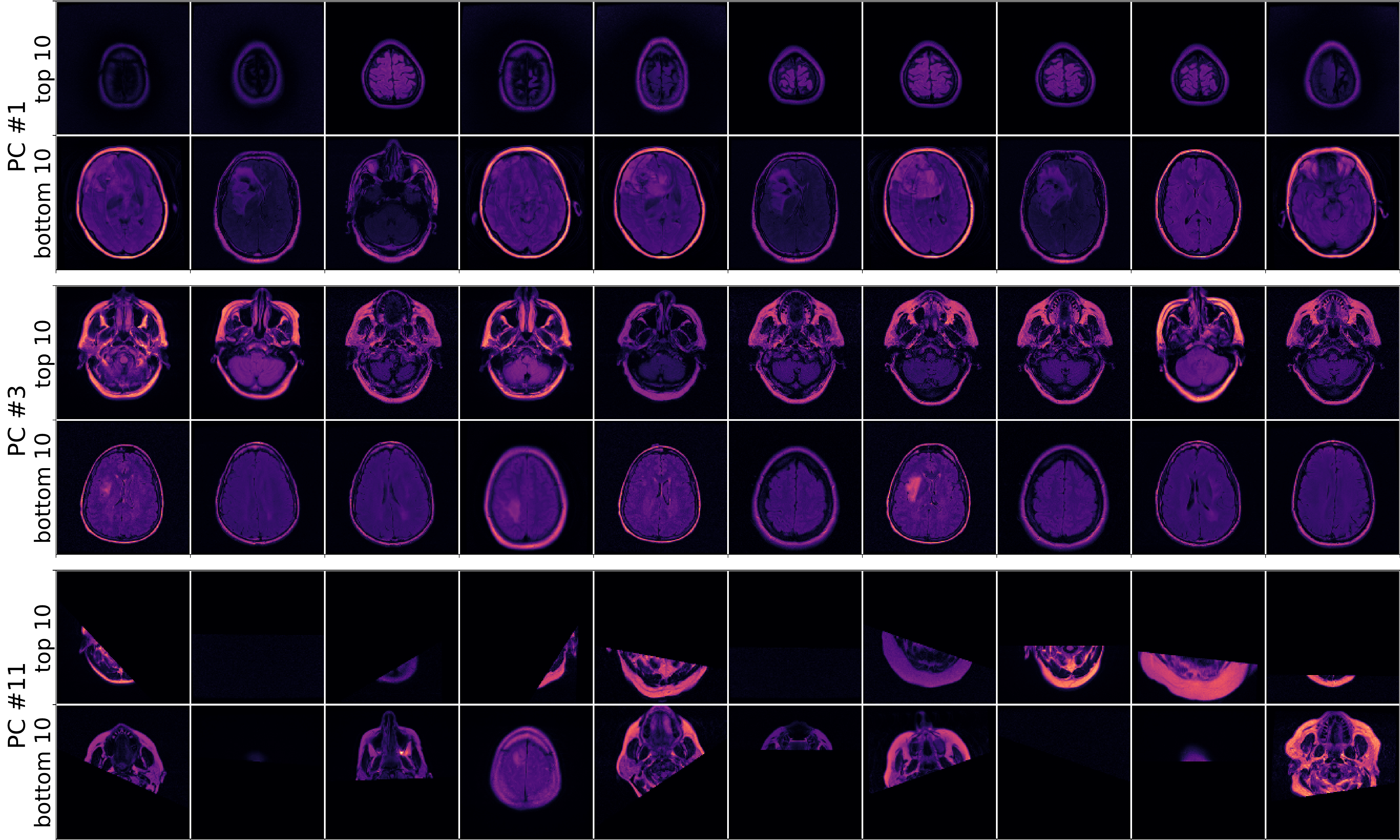}}\\
	\vspace{-5pt}
	\subfloat[CXR]{\includegraphics[width=\linewidth]{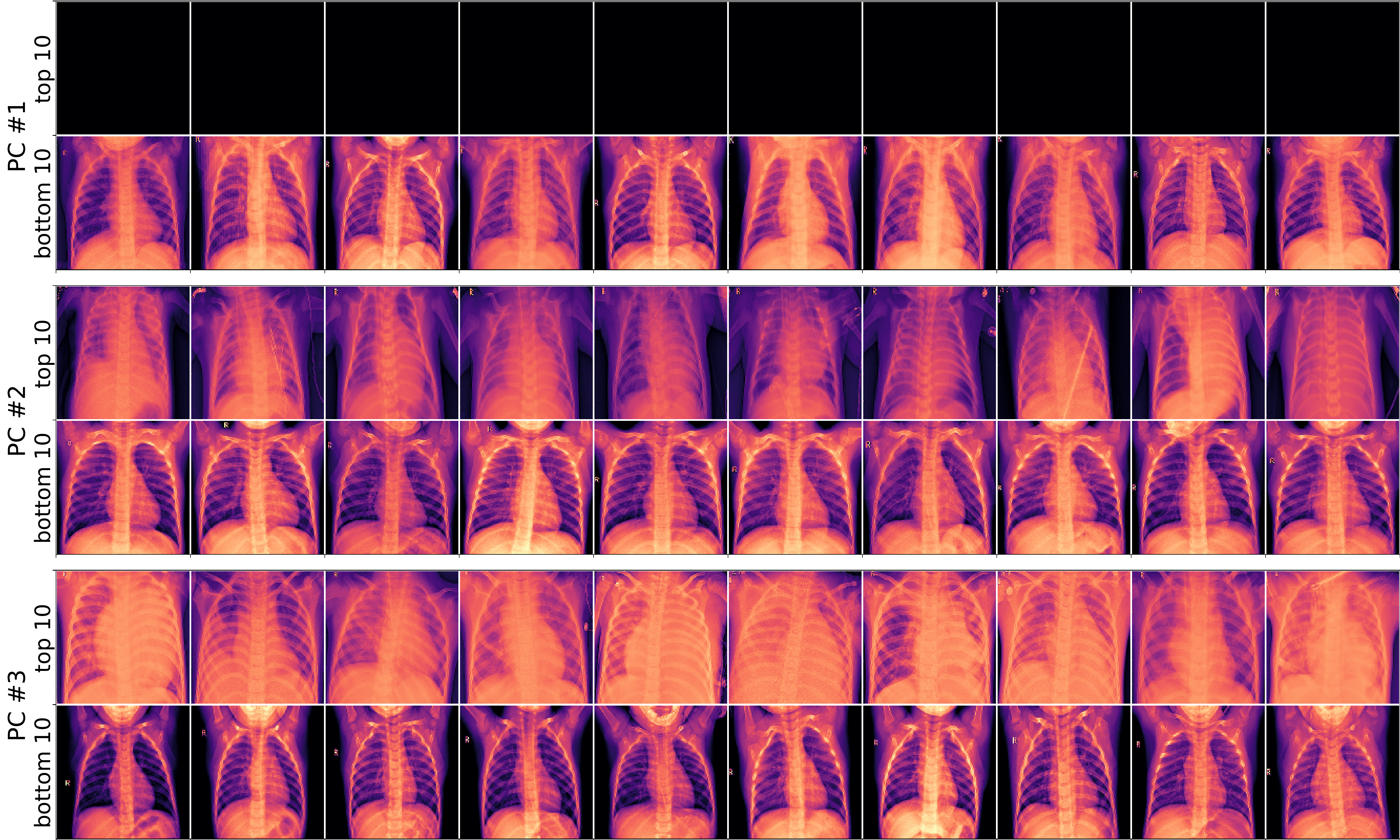}}\\
	\vspace{-5pt}
	\subfloat[OCT]{\includegraphics[width=\linewidth]{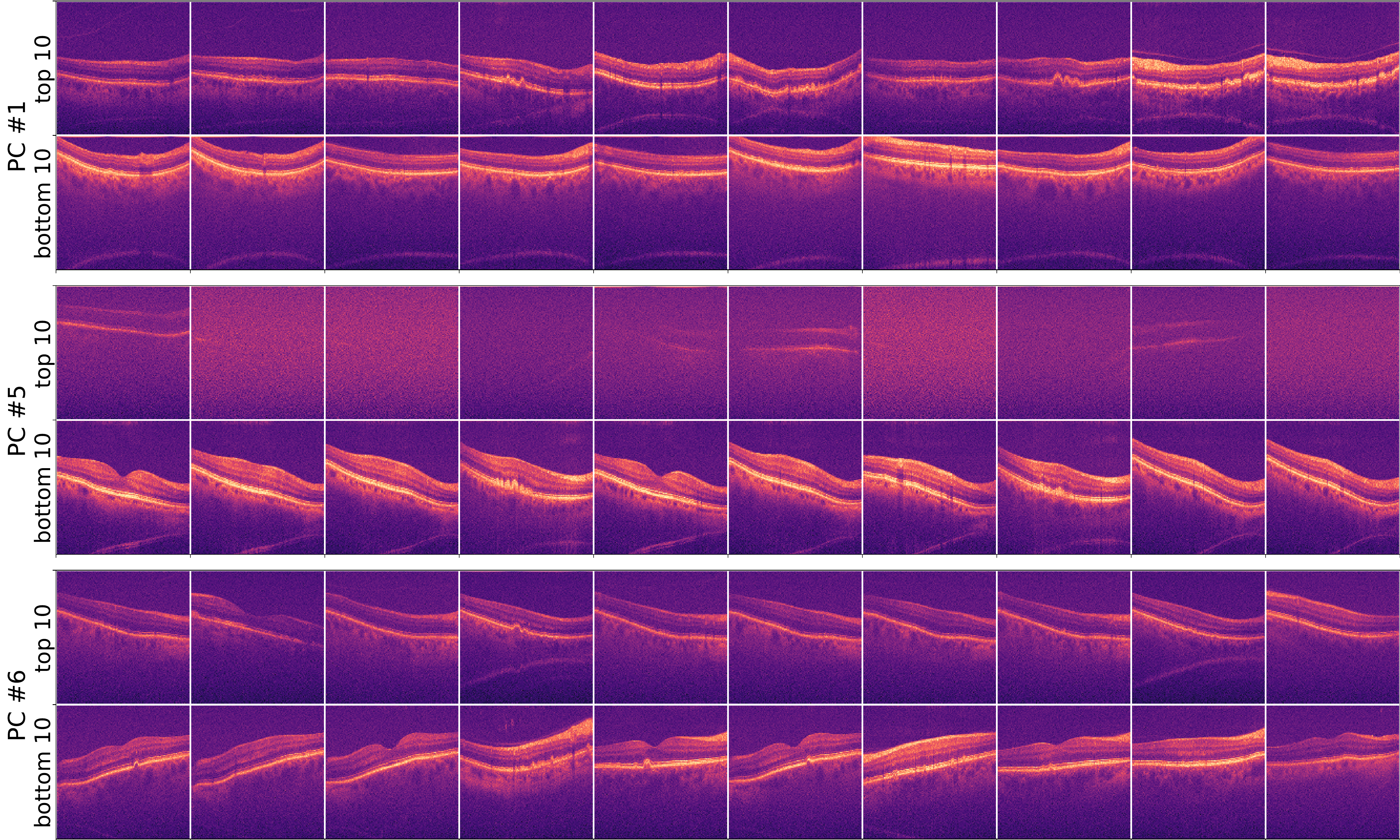}}
	\caption{Semantic groups for the MRI (a), CXR (b) and OCT(c) data. Values increase left to right. See the text for an interpretation of the groups.} \label{samples}	
\end{wrapfigure}
\section{Results}
When ordering the values for each principal component, we discovered that, for most components, the bulk of the datapoints has a value near zero, while only a fraction of samples had high positive or negative values.
The curve, which results from ordering the values, resembles a logit function to some extent (see Fig. \ref{components} in the Appendix).
Looking at samples with such extreme values, we saw that these samples, in many cases, shared some semantic structure.
In Fig. \ref{samples} we demonstrate the semantic grouping that results from our method. Each subfigure shows the results from one of the datasets.
For each dataset, we picked three principal components to visualize, showing the ten images with the highest and ten with the lowest value of that principal component.
Identifying, which principal component corresponds to which semantic concept is evident in some cases but can be difficult in others.
As such, we put the concept we think a component is responsible for in quotation marks.
Generally, we find that the difficulty of interpretation increases with the number of the principal component, as was expected given the nature of PCA.

For the MRI data, we picked the principal components for "brain size" (PC \#1), "axial position" (PC \#3) and "cutoff" (PC \#11), as shown in Fig. \ref{samples}.
The "cutoff" component reveals that some images in the dataset are cut off.
Interestingly, samples with a high value in this component are cut off at the top, while the ones with low values are cut off at the bottom.
Depending on the application, these samples should be inspected more closely and removed from the dataset if need be.

In the case of CXR data, we show the components for "non-zero" (PC \#1), "chest width" (PC \#2) and "contrast" (PC \#3).
It turns out that 270 images in the dataset have three channels (all containing the same image), while all others are grayscale.
Our I/O would read those in as float between 0 and 1 instead of uint8.
Since we also divided the samples by 255 to convert them to float, these images are almost zero.

For the OCT data, we included the components which appear to correlate with "vertical position" (PC \#1), "signal-to-noise ratio" (PC \#5) and "angle" (PC \#6).
Images with a high value in the "SNR" component show almost pure noise and are unlikely to yield useful results when used in a predictive setting.
\section{Conclusion and Future Research}
In this work, we introduced a novel method for investigating the contents of large image datasets.
Our approach aids the discovery of groups of images that share a common semantic structure.
We demonstrated the application of this method to three different medical image datasets and found relevant semantic groups in each.

However, there are still several questions, which warrant further research.
Currently, we do not know how the training duration affects the formation of semantic groups.
In particular, since our models did not overfit to the data, even when training for over a thousand epochs, we do not know if overfitting could prove beneficial in this case.
In future research, we also wish to investigate, whether regularization (e.g., sparsity) or other Autoencoder variants can help in better disentangling the semantic factors of variation.
\newpage
\bibliographystyle{ieeetr}
\bibliography{bibliography}

\begin{thebibliography}{10}

\bibitem{gordo16}
A.~Gordo, J.~Almaz{\'a}n, J.~Revaud, and D.~Larlus, ``Deep image retrieval:
  Learning global representations for image search,'' in {\em Computer Vision
  -- ECCV 2016} (B.~Leibe, J.~Matas, N.~Sebe, and M.~Welling, eds.), (Cham),
  pp.~241--257, Springer International Publishing, 2016.

\bibitem{gillmann17}
C.~Gillmann, P.~Arbeláez, J.~T. Hernández, H.~Hagen, and T.~Wischgoll,
  ``{Intuitive Error Space Exploration of Medical Image Data in Clinical Daily
  Routine},'' in {\em EuroVis 2017 - Short Papers} (B.~Kozlikova, T.~Schreck,
  and T.~Wischgoll, eds.), The Eurographics Association, 2017.

\bibitem{settles10}
B.~Settles, ``Active learning literature survey,'' tech. rep., 2010.

\bibitem{hyunha15}
H.~NAM and M.~SUGIYAMA, ``Direct density ratio estimation with convolutional
  neural networks with application in outlier detection,'' {\em IEICE
  Transactions on Information and Systems}, vol.~E98.D, no.~5, pp.~1073--1079,
  2015.

\bibitem{lakkaraju17}
H.~Lakkaraju, E.~Kamar, R.~Caruana, and E.~Horvitz, ``Identifying unknown
  unknowns in the open world: Representations and policies for guided
  exploration,'' in {\em Proceedings of the Thirty-First AAAI Conference on
  Artificial Intelligence}, AAAI'17, pp.~2124--2132, AAAI Press, 2017.

\bibitem{attenberg15}
J.~Attenberg, P.~Ipeirotis, and F.~Provost, ``Beat the machine: Challenging
  humans to find a predictive model's \&ldquo;unknown unknowns\&rdquo;,'' {\em
  J. Data and Information Quality}, vol.~6, pp.~1:1--1:17, Mar. 2015.

\bibitem{girdhar16}
R.~Girdhar, D.~F. Fouhey, M.~Rodriguez, and A.~Gupta, ``Learning a predictable
  and generative vector representation for objects,'' in {\em Computer Vision
  -- ECCV 2016} (B.~Leibe, J.~Matas, N.~Sebe, and M.~Welling, eds.), (Cham),
  pp.~484--499, Springer International Publishing, 2016.

\bibitem{hou19}
L.~Hou, V.~Nguyen, A.~B. Kanevsky, D.~Samaras, T.~M. Kurc, T.~Zhao, R.~R.
  Gupta, Y.~Gao, W.~Chen, D.~Foran, and J.~H. Saltz, ``Sparse autoencoder for
  unsupervised nucleus detection and representation in histopathology images,''
  {\em Pattern Recognition}, vol.~86, pp.~188 -- 200, 2019.

\bibitem{rumelhart86}
D.~E. Rumelhart, G.~E. Hinton, and R.~J. Williams, ``Parallel distributed
  processing: Explorations in the microstructure of cognition, vol. 1,''
  ch.~Learning Internal Representations by Error Propagation, pp.~318--362,
  Cambridge, MA, USA: MIT Press, 1986.

\bibitem{ulyanov18}
D.~Ulyanov, A.~Vedaldi, and V.~Lempitsky, ``Deep image prior,'' in {\em The
  IEEE Conference on Computer Vision and Pattern Recognition (CVPR)}, June
  2018.

\bibitem{wold87}
S.~Wold, K.~Esbensen, and P.~Geladi, ``Principal component analysis,'' {\em
  Chemometrics and Intelligent Laboratory Systems}, vol.~2, no.~1, pp.~37 --
  52, 1987.
\newblock Proceedings of the Multivariate Statistical Workshop for Geologists
  and Geochemists.

\bibitem{odena16}
A.~Odena, V.~Dumoulin, and C.~Olah, ``Deconvolution and checkerboard
  artifacts,'' {\em Distill}, 2016.

\bibitem{he16}
K.~He, X.~Zhang, S.~Ren, and J.~Sun, ``Deep residual learning for image
  recognition,'' in {\em The IEEE Conference on Computer Vision and Pattern
  Recognition (CVPR)}, June 2016.

\bibitem{farsiu14}
S.~Farsiu, S.~J. Chiu, R.~V. O'Connell, F.~A. Folgar, E.~Yuan, J.~A. Izatt, and
  C.~A. Toth, ``Quantitative classification of eyes with and without
  intermediate age-related macular degeneration using optical coherence
  tomography,'' {\em Ophthalmology}, vol.~121, no.~1, pp.~162 -- 172, 2014.

\bibitem{buda19}
M.~Buda, A.~Saha, and M.~A. Mazurowski, ``Association of genomic subtypes of
  lower-grade gliomas with shape features automatically extracted by a deep
  learning algorithm,'' {\em Computers in Biology and Medicine}, vol.~109,
  pp.~218 -- 225, 2019.

\bibitem{kermany18}
D.~S. Kermany, M.~Goldbaum, W.~Cai, C.~C.~S. Valentim, H.~Liang, S.~L. Baxter,
  A.~McKeown, G.~Yang, X.~Wu, F.~Yan, J.~Dong, M.~K. Prasadha, J.~Pei, M.~Y.~L.
  Ting, J.~Zhu, C.~Li, S.~Hewett, J.~Dong, I.~Ziyar, A.~Shi, R.~Zhang,
  L.~Zheng, R.~Hou, W.~Shi, X.~Fu, Y.~Duan, V.~A.~N. Huu, C.~Wen, E.~D. Zhang,
  C.~L. Zhang, O.~Li, X.~Wang, M.~A. Singer, X.~Sun, J.~Xu, A.~Tafreshi, M.~A.
  Lewis, H.~Xia, and K.~Zhang, ``{Identifying Medical Diagnoses and Treatable
  Diseases by Image-Based Deep Learning},'' {\em Cell}, vol.~172,
  pp.~1122--1131.e9, feb 2018.

\end{thebibliography}

\newpage
\appendix
\counterwithin{figure}{section} 
\section{Appendix}

\begin{figure}[H]
	\centering\includegraphics[width=0.7\linewidth]{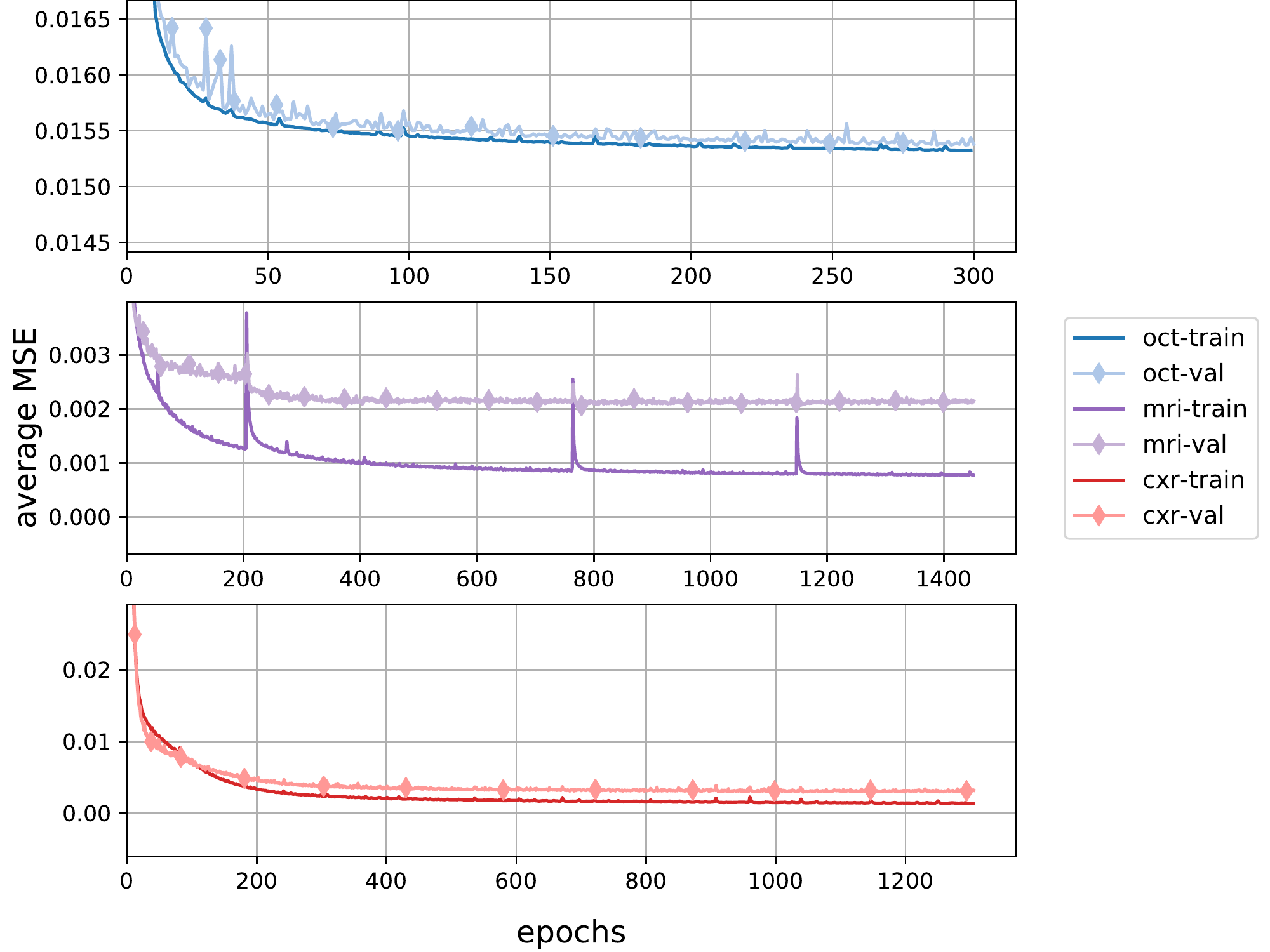}
	\caption{Training and validation losses of the Autoencoders} \label{losses}
\end{figure}
\begin{figure}[h]
	\centering
	\subfloat{\includegraphics[width=0.6\linewidth]{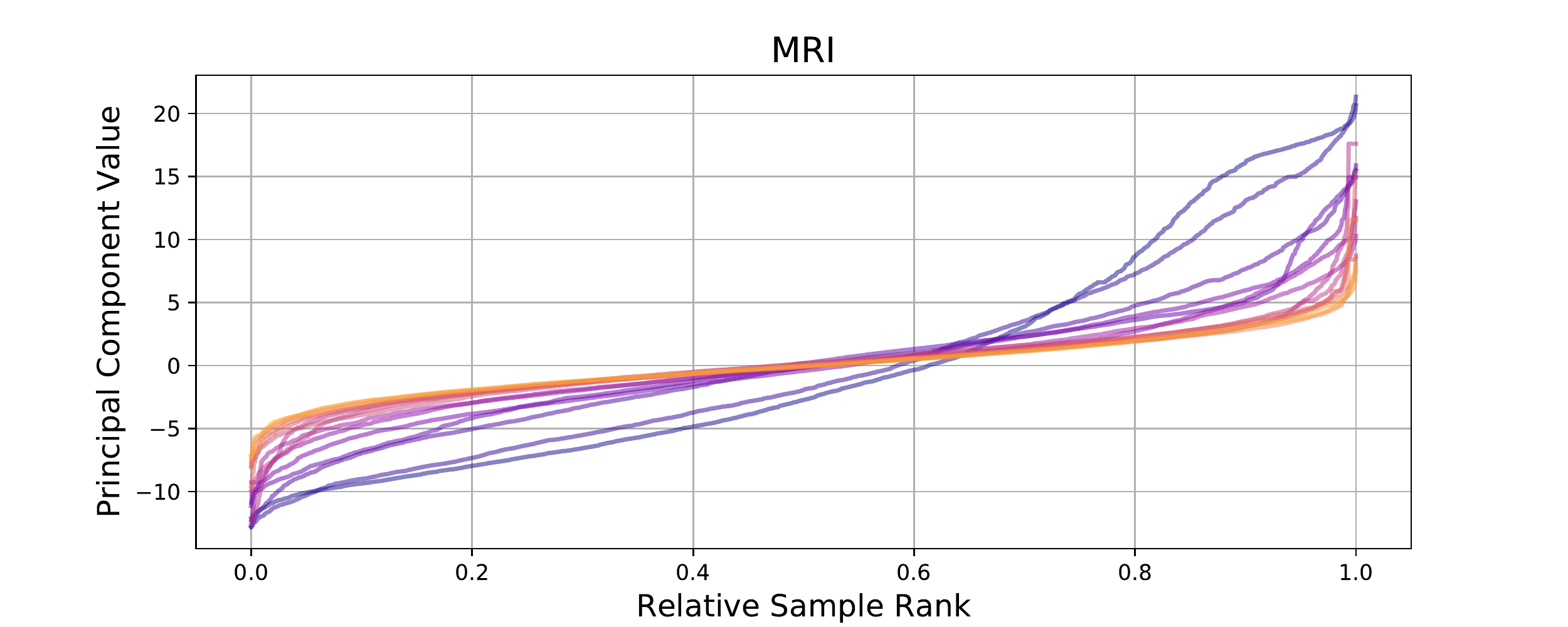}}\\
	\subfloat{\includegraphics[width=0.6\linewidth]{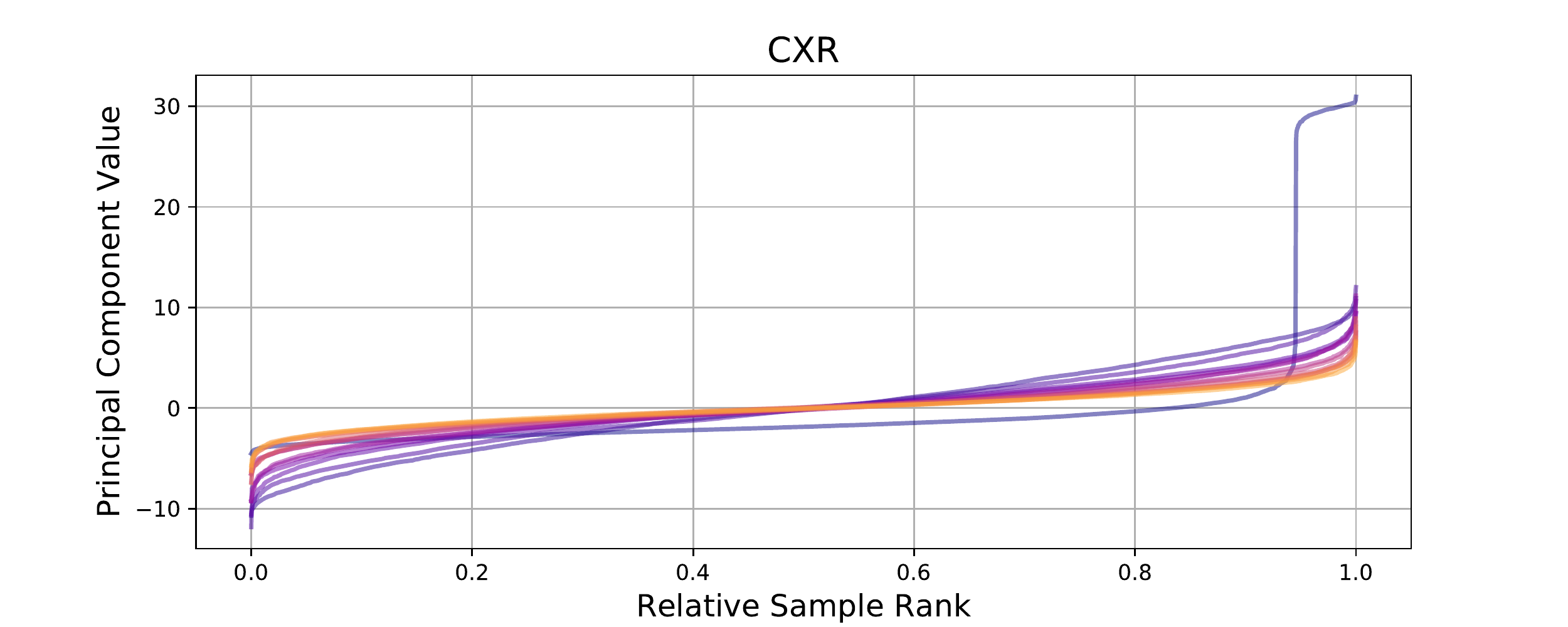}}\\		
	\subfloat{\includegraphics[width=0.6\linewidth]{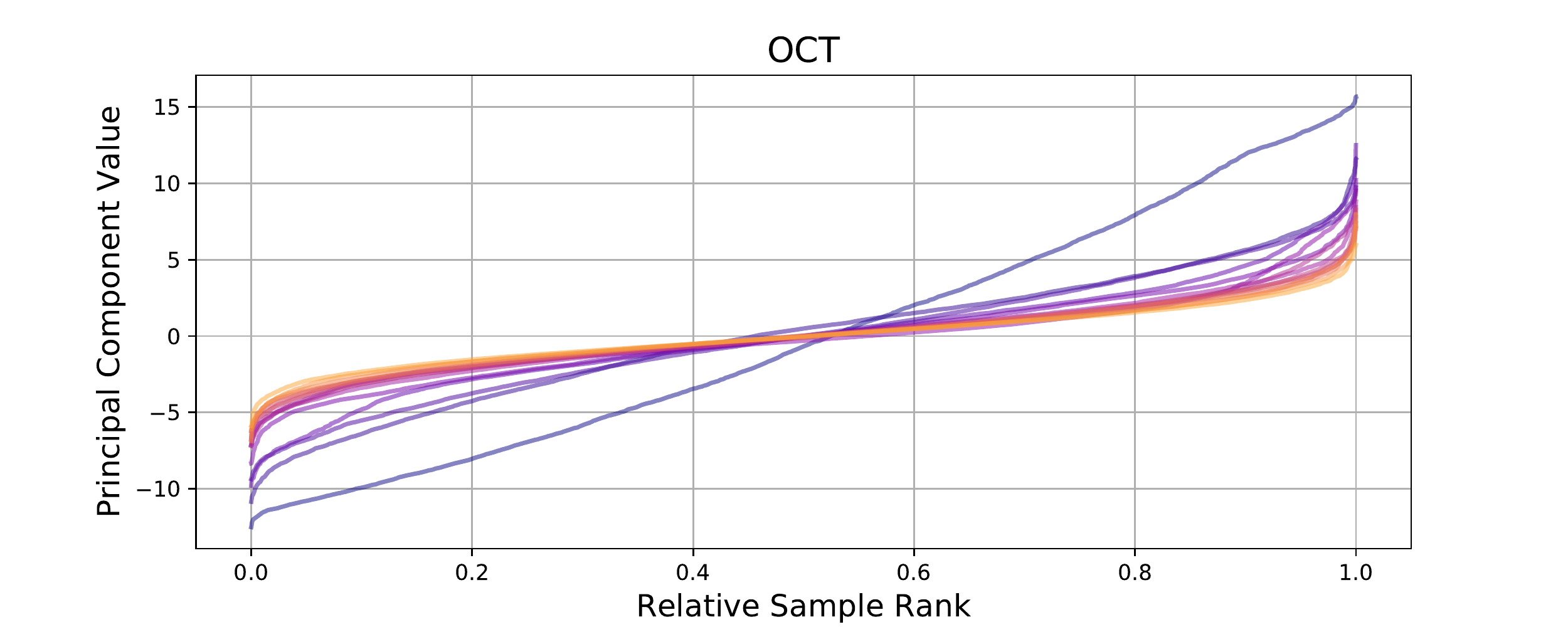}}
	\caption{Value plots for the first 15 principal components for each dataset. The darkest line represents the first principal component. Line brightness increases with the number of the principal component.} \label{components}
\end{figure}

\end{document}